\documentclass[runningheads]{llncs}
\usepackage[T1]{fontenc}
\usepackage{graphicx}
\newcommand{\mypar}[1]{\smallskip\noindent\textbf{#1.}}
\newcommand{\mypartwo}[1]{\vspace{0.5pt}\noindent\textit{#1.}}

\usepackage{amsmath}
\usepackage{amssymb}
\usepackage{amsfonts}
\usepackage{mathabx}
\usepackage{mathtools}

\usepackage{rotating}
\usepackage{varwidth}
\usepackage{float}
\usepackage{needspace}

\usepackage[hidelinks]{hyperref}
\usepackage{booktabs}
\usepackage{multirow}
\usepackage{subcaption}
\usepackage{cite}
\usepackage{longtable}
\usepackage{array}

\usepackage{algorithm}
\usepackage{algorithmic}
\usepackage{enumitem}

\usepackage{color}

\begin{document}
\title{Temporal Knowledge Graph Forecasting under Distribution Shifts: A Synthetic Evaluation}
\titlerunning{Temporal Knowledge Graph Forecasting under Distribution Shifts}
%
\author{Konrad Özdemir\inst{1} \and
Julia Gastinger\inst{1} \and \\
Lukas Kirchdorfer\inst{1,2} \and Heiner Stuckenschmidt\inst{1}}
\authorrunning{K. Özdemir et al.}
%

\institute{Data and Web Science Group, University of Mannheim, Germany \and
SAP Signavio, Walldorf, Germany}
\maketitle              
\begin{abstract}
Temporal knowledge graphs (TKGs) represent evolving relational systems, whose underlying data-generating processes often change over time. Yet, TKG forecasting models are commonly evaluated only on empirical benchmark datasets that provide limited insight into the models' robustness to such distribution shifts. Recognising this issue, we study TKG forecasting under controlled shift environments using a synthetic TKG generator that encodes three temporal and structural properties---recurrence, homophily, and periodicity---as data-generating mechanisms. This allows us to evaluate seven forecasting architectures under stationary and shifting regimes. Our experiments suggest that robustness in TKG forecasting is highly signal-dependent. Recurrence-based and periodic regularities are largely recoverable under stationary conditions, and simple memory-based baselines can be competitive when recurrence dominates the data. However, structural breaks reveal limitations in model adaptivity, with shifts in latent entity-community structure posing the strongest challenge in our study. Overall, our findings improve the understanding of the capabilities and limitations of current TKG models confronted with temporal distribution shifts.
\end{abstract}

%
%
%

\section{Introduction}
Temporal knowledge graphs (TKGs) extend static knowledge graphs (KGs) with temporal information~\cite{Han2021xerte}. As such, they constitute an effective, systematic mechanism for storing event-based facts across time. In recent years, TKG forecasting, i.e., the prediction of facts for future timestamps, has been met with rising interest. Diverse approaches have been introduced in this domain (e.g.,~\cite{REGCN,dong2025disentangled,chen2025cogntke}) and application areas may reach from finance~\cite{li2024findkg} to clinical environments~\cite{Sun2025Clinical}.

One fundamental property of such real-world temporal systems is that their underlying data-generating processes (DGPs) evolve over time, i.e., they incur shifts with respect to their underlying distribution~\cite{concept_drift_survey}. Typically, such shifts can be of sudden or gradual nature. The former may take place through the beginning of a war that can abruptly alter the interplay of diplomatic relations. The latter may arise when a change in government restructures societal systems, such as the overhaul of South Africa's education system following the end of apartheid~\cite{mouton2013}. As such, accounting for distribution shifts when modelling temporal data plays an integral role not only in classical time series forecasting, but also in adjacent sequential learning domains~\cite{Pesaran2006,KimKTPCC22,Zhang0ZLQ022}. 

In the context of single-relational temporal graph learning, recent evidence in the form of a systematic evaluation has surfaced that several approaches struggle to capture rather basic temporal patterns like periodicity~\cite{hayes2025}. Motivated, among other things, by these observations, Bl\"ocker et al.~\cite{deepgraphlearningstall2026} argue for a stronger integration of insights from network science into temporal graph learning. They highlight that decades of research on temporal and structural patterns in evolving networks still remain underutilised in modern graph learning approaches and that models are often evaluated primarily through empirical benchmark performance, without a clear understanding of the specific patterns they capture~\cite{deepgraphlearningstall2026}.

In this work, we maintain that this discrepancy extends to directed, multi-relational temporal graphs as well and, therefore, to the area of TKG forecasting. In TKG forecasting, robustness to distribution shifts remains a largely under-explored field of investigation and, to the best of our knowledge, no prior work has systematically studied how TKG forecasting architectures behave under controlled, fundamental changes in the underlying DGP. This is partly attributable to the real-data nature of most investigations; current benchmarks (e.g.,~\cite{tgb2gastinger}) do not yet provide controlled synthetic environments that enable explicit manipulation of temporal and structural properties, making it difficult to isolate which mechanisms models successfully capture and under which conditions they fail.

To address these limitations, we investigate robustness and adaptivity to distribution shifts in TKG forecasting. Building on insights from temporal graph learning, network science, and time series analysis, we contribute as follows:

\begin{enumerate}[noitemsep,topsep=0pt]
    \item We introduce a synthetic TKG data generator that produces fully controllable, temporal multi-relational graph data. The generator instantiates proxies for distribution shifts that allow for systematic investigations of model behaviour under changes in periodic patterns, entity community structure, and fact recurrence.
    \item We evaluate seven TKG forecasting approaches and evidence that robustness to distribution shifts is highly signal-dependent: on our tested datasets, recurrence-based and periodic patterns are largely recoverable under stationary conditions, but structural breaks expose limitations in adaptivity. Shifts in entity-community structure pose the strongest challenge in our study.
\end{enumerate} 

\noindent
Overall, this work aims at providing a systematic perspective on distribution shifts in TKG forecasting and to improve the understanding of the capabilities and limitations of current modelling architectures under this aspect.
\section{Related Work}\label{sec:relwork}
This section outlines common forecasting models, evaluation techniques and robustness investigations in TKG forecasting and adjacent domains.

\mypar{TKG Forecasting Architectures} 
A wide range of architectures has been proposed for future link prediction in temporal knowledge graphs. They may be classified into recurrent approaches (e.g., RE-GCN~\cite{REGCN}, CEN~\cite{li2022complex}), rule-based approaches (e.g., T-Logic~\cite{liu2022tlogic}), 
reinforcement learning variants (e.g., TimeTraveler~\cite{sun2021timetraveler}) and hybrid models (e.g., DiMNet~\cite{dong2025disentangled},  CognTKE~\cite{chen2025cogntke}).
All models are predominantly evaluated on empirical benchmark data. Consequently, relatively little is known about which temporal and structural mechanisms models successfully capture, or how robustly they generalise under evolving data conditions.

\mypar{Evaluation in TKG Forecasting}
Recent work has improved evaluation practices for TKG forecasting. A recurrence-based baseline was introduced in~\cite{gastinger2024history}, showing that simple repetition of past facts is surprisingly competitive, and that in several datasets state-of-the-art models fail to consistently outperform this baseline. In addition, the TGB 2.0 framework~\cite{tgb2gastinger} provides standardised evaluation protocols and datasets for large-scale comparisons of models.
Despite these advances, evaluation is still largely centred on aggregate benchmark performance. This provides limited insight into the temporal and structural patterns that models actually learn. In single-relational temporal graph learning, recent work has begun to address this gap by directly probing model capabilities. Hayes et al.~\cite{hayes2025} show, for example, that several methods struggle to capture rather basic temporal properties such as periodicity.
Dizaji et al.~\cite{dizaji2026} introduce a synthetic diagnostic benchmark for learning on temporal graphs and find that even on simple synthetic graphs,
models have difficulty capturing basic temporal structures.
However, analogous analyses for TKG forecasting remain missing.

\mypar{Robustness to Temporal Distribution Shifts}
Distribution shifts are well-established challenges in temporal and sequential learning, where the data distribution, or the relationship between inputs and targets, changes over time~\cite{concept_drift_survey}. These challenges have been observed across a wide range of temporal modelling domains, including recommender systems~\cite{TahmasbiJS21}, traffic forecasting~\cite{LiuSLXLBWLZ25}, and process mining~\cite{KirchdorferOAS26}, where evolving user preferences, mobility patterns, business processes have motivated domain-specific strategies for detecting and adapting to distribution shifts. In classical time-series analysis, related phenomena have long been studied under the notions of nonstationarity, with extensive work on detecting shifts and adapting approaches accordingly~\cite{Bai1998,Pesaran2006}. More recently, deep time-series forecasting has also begun to explicitly address temporal distribution shifts, for example through reversible normalisation, adaptive recurrent architectures, and nonstationary Transformer variants~\cite{KimKTPCC22,Du0FPQXW21,LiuWWL22}.
Similar concerns have emerged in graph learning~\cite{LiWZZ25} with out-of-distribution benchmarks~\cite{gui2022good} and dynamic methods to study shifts in the spatio-temporal domain explicitly~\cite{Zhang0ZLQ022}.

In TKG forecasting, however, robustness to distribution shifts has received considerably less attention. The closest related works focus on continual- or incremental TKG completion, where models are updated as new events arrive and must retain previously acquired knowledge while adapting to the evolving graph~\cite{mirtaheri_etal_2023_history}. More recent work addresses this setting through adaptive replay mechanisms to mitigate catastrophic forgetting~\cite{Zhang0LW25}. Despite considering temporal evolution and model adaptation, these studies target TKG completion rather than forecasting under controlled changes in the underlying DGP. Thus, it remains unclear which temporal and structural mechanisms current TKG forecasting models rely on, and how robust they are when these mechanisms evolve.



\mypar{This Work}
The above gaps motivate the present work. We study robustness and adaptiveness in TKG forecasting under controlled distribution shifts. Via a synthetic TKG data generation tool that enables systematic manipulation of temporal properties, we generate controlled synthetic datasets and evaluate representative TKG forecasting architectures under shifting DGP conditions.
\section{Notation and Synthetic Data Design}
This section establishes necessary notational preliminaries as well as our synthetic TKG data generation framework. 

\subsection{Notation and Background} 
Let $n_e, n_r, T\in\mathbb{N}$. Denote with $\mathcal{E}=\{0,\dots,n_e-1\}$ the \emph{entity} set and with $\mathcal{R}=\{0,\dots,n_r-1\}$ the relation set. Triplets of form $(s,r,o)$, with $s,o\in \mathcal{E}, r\in \mathcal{R}$ are referred to as \emph{facts}. Furthermore, define $\mathcal{T} = \{1,\dots, T\}$ as the set of timestamps at which facts are recorded. In our work, we focus on non-self-referential facts (cf.~\autoref{sec:self_ref_facts}) and devise the $n_e, n_r$-induced space as the \emph{universe of facts} $\Omega=\{(s,r,o): s,o\in\mathcal{E},\ r\in\mathcal{R},\ s\neq o\}$, with $\lvert\Omega\rvert = n_r (n_e^2 -n_e)$. Within this setting, our work considers TKGs as discrete graph snapshot process $(G_t)_{t\in\mathcal{T}}$ with $G_t \coloneqq \{(t, s,r,o) \mid s,o \in \mathcal{E}, r \in \mathcal{R} \}$.  Moreover, we exclude multigraph structures: \\$\forall t\in\mathcal{T}, (r, \overline{r})\in\mathcal{R}^2 \text{ s.t. } r\neq \overline{r}:(t,s,r,o)\in G_t \Rightarrow (t,s,\overline{r},o)\notin G_t$. Via $\mathbb{I}_{\textnormal{condition}}$ we denote the indicator function, equal to $1$ if `condition' holds and $0$ otherwise.

Given $(G_t)_{t\in\mathcal{T}}$, the TKG forecasting objective entails predicting future timestamped facts of the form $(t^\star, s,r,o)$ where $t^\star > T$, $r \in \mathcal{R}$, and $s,o \in \mathcal{E}$. 
In this work, we consider \emph{entity forecasting}, which focuses on predicting object or subject entities for queries of types $(t^\star, s,r,?)$ and $(t^\star, ?,r,o)$. 
This protocol reflects a common choice for assessing the capabilities of TKG forecasting methods~\cite{gastinger2023eval}. 
Typically, TKG forecasting is formulated as a ranking problem~\cite{han2022TKG}: Given a query derived from a fact in the test or validation set, a model assigns plausibility scores to candidate entities in $\mathcal{E}$ and ranks them accordingly. 

\subsection{Synthetic Data Design}
To study model behaviour under controlled distributional structure, we define synthetic TKGs as time-indexed sampling processes over the universe of admissible facts $\Omega$. Our experimental design comprises three ablations, each isolating a distinct temporal-structural mechanism: \emph{recurrence}, \emph{periodicity}, and \emph{homophily}. Fact recurrence is a well-documented property of empirical TKG datasets~\cite{gastinger2024history}. We therefore examine how well models exploit recurrent facts when this signal is isolated from other confounding factors. Periodicity captures seasonal regularities in temporal data and is a central concept in time series analysis~\cite{box2015time}. In temporal networks, such patterns arise in diverse settings, e.g., in social interaction data. Yet, they remain comparatively underexplored in temporal graph modelling~\cite{qin2019mining}. We therefore include a periodicity variant to assess whether the models can recover seasonal fact patterns. Finally, we consider homophily, a fundamental mechanism in network science. Many real-world networks exhibit community structure, where entities interact preferentially with members of the same or related groups~\cite{networksNewman}. The homophily variant allows us to evaluate whether the models can capture such group-based interaction regularities. Importantly, for the above three variants, we investigate a \emph{persistent}- and a \emph{break} setting, respectively. The former aims at discovering whether prospective models are able to capture the variant-related signal, while the latter aims at discovering whether they are capable of adapting to a sudden distribution shift in the dataset. 

For each timestamp $t\in\mathcal{T}$, let $m_t\in\mathbb{N}$ denote the snapshot edge budget, with $m_t\leq |\Omega|$. A snapshot $G_t$ is generated by sampling $m_t$ distinct triples from $\Omega$ and recording them as timestamped facts $(t,s,r,o)$. All variants can be described by a sampling weight function $w_t$. The 
probability mass function is
\begin{equation*} \label{eq:base-sampling-pt}
p_t(s,r,o)=\frac{w_t(s,r,o)}
{\sum_{(s',r',o')\in\Omega} w_t(s',r',o')}.
\end{equation*}
An observed snapshot is then obtained by sampling $p_t$ without replacement. The break setting is instantiated via a \emph{break date} $\tau^\star$, which indicates the first timestamp where the distribution shift comes to effect.

\mypar{Null}
The null variant serves as the negative-control process. At each timestamp, all admissible triples receive identical weight
\begin{equation*}
    w_t(s,r,o)=k, \quad k>0.
\end{equation*}
Consequently, each snapshot is sampled uniformly from $\Omega$, subject only to the fixed edge budget. 
Note that systematic performance above a random-ranking reference on data of this variant cannot be attributable to a meaningful signal in the DGP, but rather to, e.g., finite-sample effects or evaluation-set structure.

\mypar{Recurrence}
The recurrence variant introduces dependence on previously observed facts. For a triple $x = (s,r,o)\in\Omega$, the \emph{recurrence state} before timestamp~$t$ is represented by a history score with exponential decay parameter $\delta\in(0,1]$: $h_t(x) = \sum_{\tau<t}
\delta^{t-\tau}\mathbb{I}_{\{(\tau,s,r,o)\in G_\tau\}}$. The sampling weight is then given by
\begin{equation*}
w_t(x)=\exp\left(\rho h_t(x)\right),\qquad \rho\geq 0,
\end{equation*}
where $\rho$ controls the strength of the recurrence signal. As such, triples that have appeared previously receive increased sampling probability in future snapshots.

In the persistent variant, the recurrence state evolves only through the accumulation and decay of past observations. In the break variant, the recurrence strength may change, i.e., $\rho_{pre}$ can differ from $\rho_{post}$. Moreover, the memory state may be partially reset at the break date $\tau^\star$: let $B(x)\sim \textnormal{Ber}(p)$ for $p\in[0,1]$. Then, once at the break timestamp, the history is modified via $h_{\tau^\star}(x)
=B(x)h_{\tau^\star - 1}(x), \ x\in\Omega$. Facts where $B=0$ lose their accumulated recurrence score, while facts with $B=1$ retain it. After the reset, the recurrence state again evolves according to the same update rule. This variant tests whether forecasting models exploit persistence in fact occurrence and whether they adapt when part of the historical recurrence information is lost.

\mypar{Homophily}
The homophily variant introduces community structures in the entity space $\mathcal{E}$. Let $K\in\mathbb{N}$ denote the number of latent classes and let
$c_t:\mathcal{E}\rightarrow{1,\dots,K}$ assign each entity to a class at timestamp $t$. The sampling weight of a triple then depends on whether its subject and object belong to the same  class:
\begin{equation*}
    w_t(s,r,o)=\exp\left(
    \gamma \ \mathbb{I}_{\{c_t(s)=c_t(o)\}}
    \right),
    \qquad
    \gamma\geq 0 .
\end{equation*}
For $\gamma=0$, the process reduces to uniform sampling over $\Omega$. For $\gamma>0$, triples connecting entities in the same latent class receive larger sampling weight than cross-class triples. Relations are not class-specific in this variant; the signal is carried by the induced block structures over entities.

In the persistent setting, the class assignment is time-invariant, i.e., $c_t(e)=c(e)$, $e\in\mathcal{E},\ t\in\mathcal{T}$.
In the break setting, the homophily strength $\gamma$ may change from $\gamma_{pre}$ to $\gamma_{post}$, and the latent class allocation may also do so at $\tau^\star$. In that sense, there are two class maps $c^{\mathrm{pre}}$ and $c^{\mathrm{post}}$ such that $c_t=c^\mathrm{pre}\mathbb{I}_{\{t < \tau^\star\}} + c^\mathrm{post}\mathbb{I}_{\{t \geq \tau^\star\}}$. The number of classes remains $K$, but a part of the partition of $\mathcal{E}$ is reorganised at the shift, which enables the effect of a changing entity community structure. Notably, extreme choices of $K$ recover the null data. If $K=1$, all admissible triples connect entities within the same class, so all triples receive the same weight $\exp(\gamma)$. If $K=n_e$ and each entity forms its own class, then $s\neq o$ implies $c_t(s)\neq c_t(o)$, so all admissible triples receive weight $1$. 

\mypar{Periodicity}
The periodicity variant introduces periodical, time-dependent variations in the fact-sampling behaviour for each snapshot. Let $\mathcal{L}=\{H,L,N\}$ denote a finite set of temporal signal labels, where $H$ and $L$ correspond to restricted signal supports and $N$ denotes the null support. For $\ell\in\mathcal{L}$, let $\mathcal{E}_\ell\subseteq\mathcal{E}$ and $\mathcal{R}_\ell\subseteq\mathcal{R}$ define the entity and relation subsets associated with signal label~$\ell$. The corresponding support is $\Omega_\ell=\{(s,r,o)\in \mathcal{E}_\ell\times \mathcal{R}_\ell\times \mathcal{E}_\ell : s\neq o\}$.
The null label uses the full admissible universe, $\Omega_N=\Omega$. Denoting with $L^q = \bigtimes_{i=1}^q \mathcal{L}$ the $q$-fold cartesian product of the label set, 
temporal variation is induced by a sequence $\pi\in\mathcal{L}^q$. At a timestamp $t$, the assigned label is $\ell_t=\pi_{1+((t-1)\bmod q)}$, and the corresponding sampling space $\Omega_{\ell_t}$. The sampling weights are given by 
\begin{equation*}
    w_t(s,r,o)=\mathbb{I}_{\{(s,r,o)\in\Omega_{\ell_t}\}}.
\end{equation*} 
Thus, conditional on the active temporal phase, all admissible triples inside $\Omega_{\ell_t}$ are equiprobable, whereas triples outside this space cannot be sampled.

\begin{figure}
    \centering
    \includegraphics[width=0.8\linewidth]{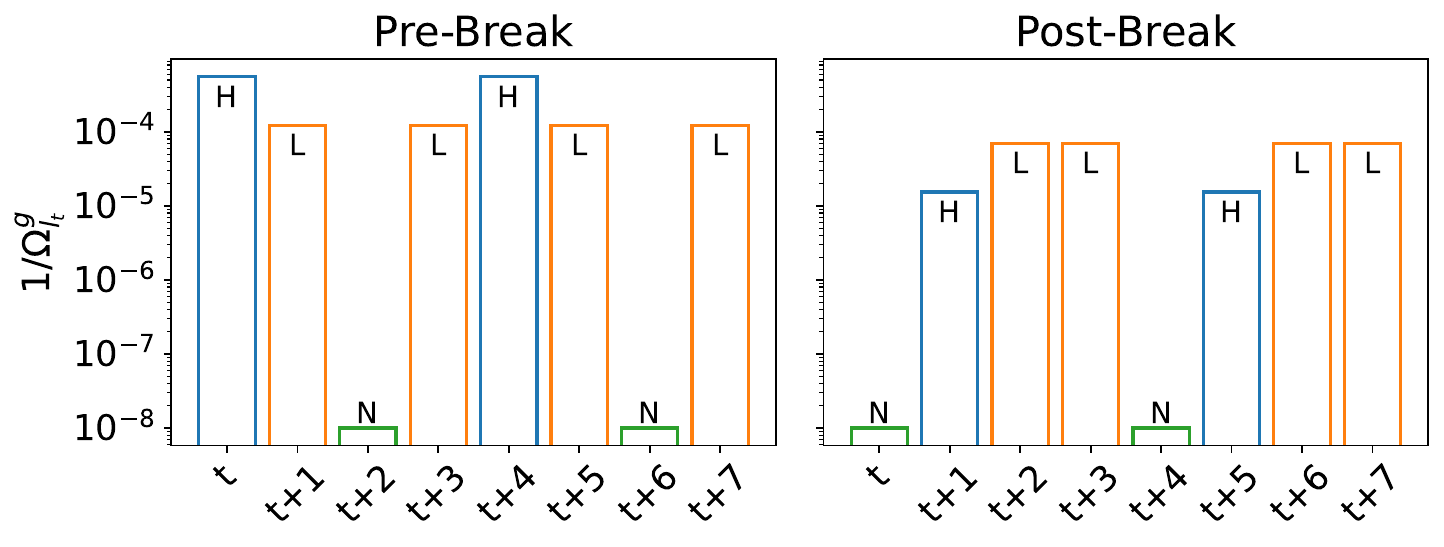}
    \caption{Schematical periodic pattern of three signal types, measured by inverse subspace-sampling size.}
    \label{fig:period}
    \vspace{-1cm}
\end{figure}

The break setting is established as follows: the periodic mechanism changes at a shift timestamp $\tau^\star\in\mathcal{T}$ from one regime into another. Specifically, for the break setting, a regime $g$ can admit two states: $g\in\{\mathrm{pre},\mathrm{post}\}$. Each regime has its own periodic label sequence $\pi^g=(\pi^g_1,\dots,\pi^g_{q_g})$, where $q_g$ denotes the regime's period length. The active regime at timestamp $t$ is $g(t)=\mathrm{pre}*\mathbb{I}_{\{t < \tau^\star\}} + \mathrm{post}*\mathbb{I}_{\{t \geq \tau^\star\}}$. Within the active regime, the label is obtained by cycling through the corresponding periodic sequence $\ell_t=\pi^{g(t)}_{1+((t-1)\bmod q_{g(t)})}$. The resulting sampling space at timestamp $t$ is then given by $\Omega_{\ell_t}^{g(t)}$. Thus, the break may change both the periodic pattern and the signal induced by the sampling space shift. The pre- and post-break sampling spaces may differ in size and composition with the following caveat: we impose a nesting relation between the corresponding pre- and post-break supports when the induced sub-space sizes differ. This means that $\Omega_{\ell}^{\textnormal{pre}} \subseteq \Omega_{\ell}^{post}$ if $\lvert\Omega_{\ell}^{\textnormal{pre}}\rvert \leq \lvert\Omega_{\ell}^{\textnormal{post}}\rvert$ and vice-versa. This makes the break interpretable as a controlled expansion or contraction of the signal subspace. An example pattern is presented in~\autoref{fig:period}, where $q=4$ for all regimes and $\pi^{\textrm{pre}} = (H,L,N,L)$ and $\pi^{\textrm{post}} = (N,H,L,L)$ with both sub-spaces $H,L$ expanding post-break.

\vspace{-0.3cm}
\section{Modelling Setup and Evaluation}
This section defines the setup used to assess TKG forecasting methods on the proposed synthetic datasets. We first introduce the common dataset- and variant-specific parameters used in the experiments as well as the evaluation protocol. We note that the chosen parameters constitute exemplary configurations and may, up to the interested reader's preferences, be altered to varying degrees. Finally, we specify the link-prediction architectures and baselines included in the comparison before defining an oracle reference model that scores candidates using the known data-generating process. 
The Code is available on GitHub\footnote{ \url{https://github.com/konradoezdemir/shifting-tkg}, framework and experiment setup.}.


\subsection{Data- and Evaluation Protocol}
All synthetic datasets use $n_e=2500$ entities, $n_r=16$ relations, and $T=100$ timestamps. Excluding self-referential facts yields $|\Omega|=9996\times 10^4$ admissible facts. Also, we devise an \emph{observation ratio} of $10^{-4}$ (cf.~\autoref{sec:observation_ratio} for details), such that each dataset contains $9996$ facts in total, which we simplify to $m_t=100$ facts per snapshot (i.e., our edge budget). We fix break events at timestamp $\tau^\star=67$ to ensure that the models have seen some (3), but not many events after $\tau^\star$ during training. Subsequent hyperparameters are elaborated in~\autoref{ap:hyperparam_choices_synTKG}.

\mypar{Recurrence}
The recurrence variants deploy a decay parameter of $\delta=0.95$ and recurrence strength $\rho=7.5$ for both settings. In the break setting, at $\tau^{\star}$, $p=50\%$ of the accumulated recurrence memory is reset. 

\mypar{Homophily}
The homophily variants use $K=250$ latent entity classes, corresponding to an average class size of $10$. The homophily strength is fixed at $\gamma=7.5$ for both settings. At break $\tau^\star$, $50\%$ of the class structure is reorganised and, in our experiment, approx. $42.31\%$ of pre-break facts change their implied status (same-class to cross-class and cross-class to same-class).

\mypar{Periodicity}
The periodicity variant uses labels $H$, $L$, and $N$, where $H$ and $L$ denote restricted signal supports and $N$ represents the full universe $\Omega$. The persistent setting follows the pattern $\pi =(H,L,N,L)$. Here, $H$ support entails $n_e^H = 30$ entities and $n_r^H =2$ relations, while $L$ entails $n_e^L = 45$ and $n_r^L =4$. The break setting initially follows the same pattern and support configuration as the persistent one. At $\tau^{\star}$, the pattern changes to $\pi^{\textnormal{post}}=(N,H,L,L)$, preserving the frequency of each label while altering their temporal ordering. Simultaneously, the support sizes are expanded: the $H$ subspace increases to $n_e^H =90$ and $n_r^H =8$ and the $L$ subspace to $n_e^L =60$ and $n_r^L =4$. Figure~\ref{fig:period} illustrates the patterns.

\mypar{Evaluation Protocol}
We evaluate model performance in a single-step setting, closely following the protocol outlined in TGB 2.0~\cite{tgb2gastinger}. We split each dataset into a train-, validation-, and test set with a $70:15:15$ ratio.  Following~\cite{tgb2gastinger}, the dataset preprocessing adds inverse edges so that both tail and head prediction can be evaluated in a tail-prediction format. 
Overall, we report our findings (mean and standard deviation across five runs) via the time-aware Mean Reciprocal Rank (MRR) and the proportion of true facts correctly assigned within the first $k\in\mathbb{N}$ positions (Hits@k). 
However, tie handling differs from TGB 2.0: When multiple entities receive the same score, we do not assign mean ranks. Instead, we calculate the expected rank of all tied scores given a uniform sampling approach~\cite{sun2020re}. The chief reason lies in the nature of our oracle (cf.~\autoref{subsec:oracle}) and the associated high likelihood of many candidate entities being scored equally, yielding occurrences of large, grouped ties. For models producing a continuous score (e.g., RE-GCN), this should de-facto equate the mean-rank tie handle.

\subsection{Link Prediction Architectures}
In our experiments, we compare representative methods spanning the major classes of TKG forecasting architectures that we described in Section~\ref{sec:relwork}\footnote{Please find information on all hyperparameters in Appendix~\ref{ap:models}.}:
RE-GCN~\cite{REGCN}, TLogic~\cite{liu2022tlogic}, TimeTraveler~\cite{sun2021timetraveler}, DiMNet~\cite{dong2025disentangled}, and CognTKE~\cite{chen2025cogntke}, as well as two heuristic baselines: The single-relational EdgeBank~\cite{poursafaei2022towards}, which we modify to the multi-relational setting by incorporating relation types into stored facts; we refer to this variant as \emph{EdgeBank-RX}. In addition, we include the Recurrency Baseline~\cite{gastinger2024history}. Finally, we report random guessing as a lower-bound, and our oracle scorer as an upper-bound reference, which we detail below.


\subsection{Oracle Reference} \label{subsec:oracle}
We introduce an oracle scorer that enables comparisons across datasets, even when they differ in test sets or underlying sampling distributions.
This reference model \emph{knows} the DGP and does therefore not learn from data. 
Specifically, it has access to the probabilities that were used to generate the data, but does not know which edge was sampled at a particular timestamp, except insofar as this is encoded in the past when the generator itself is history-dependent.
In doing so, it represents an upper bound of capturable signal \emph{in distribution}, i.e., over many runs. 
Concretely, for each tail-prediction query $(t,s,r,?)$, the oracle scores 
all candidates using the true data variant generation weights at timestamp $t$. 
For a candidate object $o$, it constructs $x_o=(s,r,o)$ and assigns the score 
\begin{equation*}
    \mathrm{score}_t(x_o)=\log w_t(x_o)-\log \sum_{x'\in\Omega} w_t(x'),
\end{equation*}
where the logarithm is used to expand the scoring range, although any strictly monotonic transformation would preserve the ranking.

\section{Results}
This section presents our experimental results, assessing the TKG forecasting models across all three data variants in both persistent and break settings.

\mypar{Recurrence}
For the recurrence variant (cf.~\autoref{tab:recurrence-results}), the persistent setting demonstrates that the injected signal is recoverable by most models. Specifically, all models except DiMNet reach at least 90\% of the oracle reference, indicating that the persistent recurrence mechanism is well aligned with models capable of exploiting repeated historical facts, including simple memory-based methods.

\vspace{-0.6cm}
\begin{table}[h]
\centering
\caption{Recurrence results: MRRs are mean (s.d.) over five runs, in percent.}
\label{tab:recurrence-results}
\scriptsize
\setlength{\tabcolsep}{4pt}
\begin{tabular}{lrrrr}
\toprule
& \multicolumn{2}{c}{Persistent} & \multicolumn{2}{c}{Break} \\
\cmidrule(lr){2-3}\cmidrule(lr){4-5}
Model & MRR (s.d.) & \%Oracle MRR & MRR (s.d.) & \%Oracle MRR \\
\midrule
\textsc{Oracle} & 20.25 (0.00) & 100.00 & 16.70 (0.00) & 100.00 \\
Random & 0.34 (0.00) & 1.68 & 0.34 (0.00) & 2.04 \\
\midrule
CognTKE & 20.23 (0.03) & 99.89 & 16.53 (0.23) & 98.94 \\
DiMNet & 15.95 (0.05) & 78.77 & 12.20 (0.41) & 73.04 \\
EdgeBank-RX & 19.04 (0.00) & 94.03 & 15.80 (0.00) & 94.59 \\
RecurrencyBaseline & 20.38 (0.00) & 100.62 & 16.35 (0.00) & 97.87 \\
RE-GCN & 18.40 (0.31) & 90.87 & 13.90 (0.78) & 83.20 \\
TimeTraveler & 19.61 (0.59) & 96.82 & 15.71 (0.36) & 94.08 \\
TLogic & 18.95 (0.27) & 93.57 & 12.94 (0.68) & 77.49 \\
\bottomrule
\end{tabular}
\end{table}

\vspace{-0.3cm}

The break setting changes this picture. Since the oracle reference itself decreases from 20.25 to 16.70 MRR, absolute MRR drops should not be interpreted directly as losses in relative signal recovery. Oracle-normalised scores show that CognTKE, the Recurrency Baseline, EdgeBank-RX, and TimeTraveler remain close to the oracle reference, achieving comparable performance as in the persistent setting.
In contrast, particularly the performances of RE-GCN and TLogic drop substantially. As a result, the recurrence break primarily distinguishes methods that appear more robust to changes in the recurrence set from those whose temporal representations or rule structures appear less adaptive.

The strong performance of both recurrence-based baselines\footnote{Notably, the Recurrency Baseline slightly exceeds the oracle reference in the persistent setting: given the sampling-based nature of the underlying prediction task, such outcomes may occasionally occur. Oracle scores should therefore not be interpreted as strict upper bounds, but rather as upper bounds in distribution.} (EdgeBank-RX and RecurrencyBaseline) further illustrates that, in recurrence-dominated data, simple repetition of facts can be highly competitive with more complex neural architectures; this finding is consistent with insights from related work~\cite{gastinger2024history}.



\mypar{Homophily}
For the homophily variant (cf.~\autoref{tab:homophily-results}), the persistent setting exhibits a different model hierarchy. The best-performing methods are DiMNet, TimeTraveler, and CognTKE, reaching oracle reference scores of up to 88\%. While RE-GCN and TLogic show mediocre performances, the two recurrency-based methods clearly fall behind with scores of only 4.8\% and 6.2\%. This suggests that the homophily signal cannot be effectively captured through recurrence-based mechanisms and instead appears to require models to learn the latent community structure underlying the graph.

\vspace{-0.6cm}
\begin{table}[h]
\centering
\caption{Homophily results. MRRs are mean (s.d.) over five runs, in percent.}
\label{tab:homophily-results}
\scriptsize
\setlength{\tabcolsep}{4pt}
\begin{tabular}{lrrrr}
\toprule
& \multicolumn{2}{c}{Persistent} & \multicolumn{2}{c}{Break} \\
\cmidrule(lr){2-3}\cmidrule(lr){4-5}
Model & MRR (s.d.) & \%Oracle MRR & MRR (s.d.) & \%Oracle MRR \\
\midrule
\textsc{Oracle} & 27.37 (0.00) & 100.00 & 27.27 (0.00) & 100.00 \\
Random & 0.34 (0.00) & 1.24 & 0.34 (0.00) & 1.25 \\
\midrule
CognTKE & 21.55 (0.28) & 78.73 & 16.02 (0.23) & 58.77 \\
DiMNet & 23.99 (0.95) & 87.65 & 12.97 (0.50) & 47.55 \\
EdgeBank-RX & 1.71 (0.00) & 6.23 & 1.69 (0.00) & 6.18 \\
RecurrencyBaseline & 1.32 (0.00) & 4.83 & 1.35 (0.00) & 4.96 \\
RE-GCN & 17.57 (0.35) & 64.20 & 9.86 (0.12) & 36.16 \\
TimeTraveler & 22.83 (0.73) & 83.42 & 14.89 (1.06) & 54.60 \\
TLogic & 10.81 (0.50) & 39.50 & 6.20 (0.29) & 22.75 \\
\bottomrule
\end{tabular}
\end{table}
\vspace{-0.3cm}


In the break setting, the oracle reference remains nearly unchanged, whereas model performances degrade substantially.
Thus, compared to the persistent setting, the homophily break disproportionately affects methods relying on stable structural or temporal assumptions.
For example, DiMNet drops from 88\% of oracle MRR in the persistent setting to only 48\% after the break.
We conjecture the following properties to hold accountability for this observation: DiMNet explicitly models a stable factor intended to capture the node evolution without deviating from the node’s steady-state characteristics and intrinsic properties in the data~\cite{dong2025disentangled}. Moreover, a disentangle-component is included to separate active and stable features. While the stable features seem beneficial in the persistent setting, the break (partially) obstructs this condition of stability, likely reducing the alignment between the learned stable factor and the true underlying data distribution and thus leading to degraded performance.

In contrast, CognTKE shows comparatively strong robustness, likely due to this particular model design choice: For any given query, instead of relying on fixed entity embeddings, CognTKE dynamically constructs temporal reasoning graphs~\cite{chen2025cogntke}. Nevertheless, its remaining performance drop indicates that its path retrieval mechanism and reasoning components still depend on historical relational patterns which are disrupted at break time $\tau^{\star}$.

Overall, the homophily break setting represents a challenging structural shift in which no model preserves more than a moderate fraction of the oracle signal, highlighting limited robustness to changes in entity-class assignments.

\mypar{Periodicity}
For the periodicity variant (cf.~\autoref{tab:periodicity-results}), the persistent setting is highly accessible to most models. CognTKE, DiMNet, the Recurrency Baseline, and RE-GCN all achieve near-oracle performance. The remaining three methods fall behind, yet still attain at least 84\%.

\vspace{-0.6cm}
\begin{table}[h]
\centering
\caption{Periodicity results. MRRs are mean (s.d.) over five runs, in percent.}
\label{tab:periodicity-results}
\scriptsize
\setlength{\tabcolsep}{4pt}
\begin{tabular}{lrrrr}
\toprule
& \multicolumn{2}{c}{Persistent} & \multicolumn{2}{c}{Break} \\
\cmidrule(lr){2-3}\cmidrule(lr){4-5}
Model & MRR (s.d.) & \%Oracle MRR & MRR (s.d.) & \%Oracle MRR \\
\midrule
\textsc{Oracle} & 8.30 (0.00) & 100.00 & 5.83 (0.00) & 100.00 \\
Random & 0.34 (0.00) & 4.10 & 0.34 (0.00) & 5.83 \\
\midrule
CognTKE & 8.24 (0.26) & 99.30 & 5.36 (0.19) & 91.96 \\
DiMNet & 8.20 (0.14) & 98.77 & 4.85 (0.18) & 83.24 \\
EdgeBank-RX & 7.04 (0.00) & 84.74 & 3.19 (0.00) & 54.81 \\
RecurrencyBaseline & 8.28 (0.00) & 99.77 & 5.05 (0.00) & 86.69 \\
RE-GCN & 8.25 (0.26) & 99.34 & 5.23 (0.29) & 89.76 \\
TimeTraveler & 7.59 (0.28) & 91.46 & 5.17 (0.20) & 88.71 \\
TLogic & 7.72 (0.21) & 93.00 & 4.43 (0.12) & 75.96 \\
\bottomrule
\end{tabular}
\end{table}
\vspace{-0.3cm}

This indicates that the persistent periodic signal is comparatively easy to exploit, with model scores similar to those for the recurrence variant. We conjecture this to be due to the fact that periodicity, in this design, can be roughly identified as a combination of recurrence and homophily: the seasonal pattern induces stable and repeatedly observable temporal regularities, and the disjoint sampling spaces induce community structure between entities. Under this presupposition and the observation that models that score well on the recurrency variant tend to do so for periodicity, too, it is plausible that the recurrence side is more dominant than the homophily side. 

\begin{figure}
\centering
\includegraphics[width=1\linewidth]{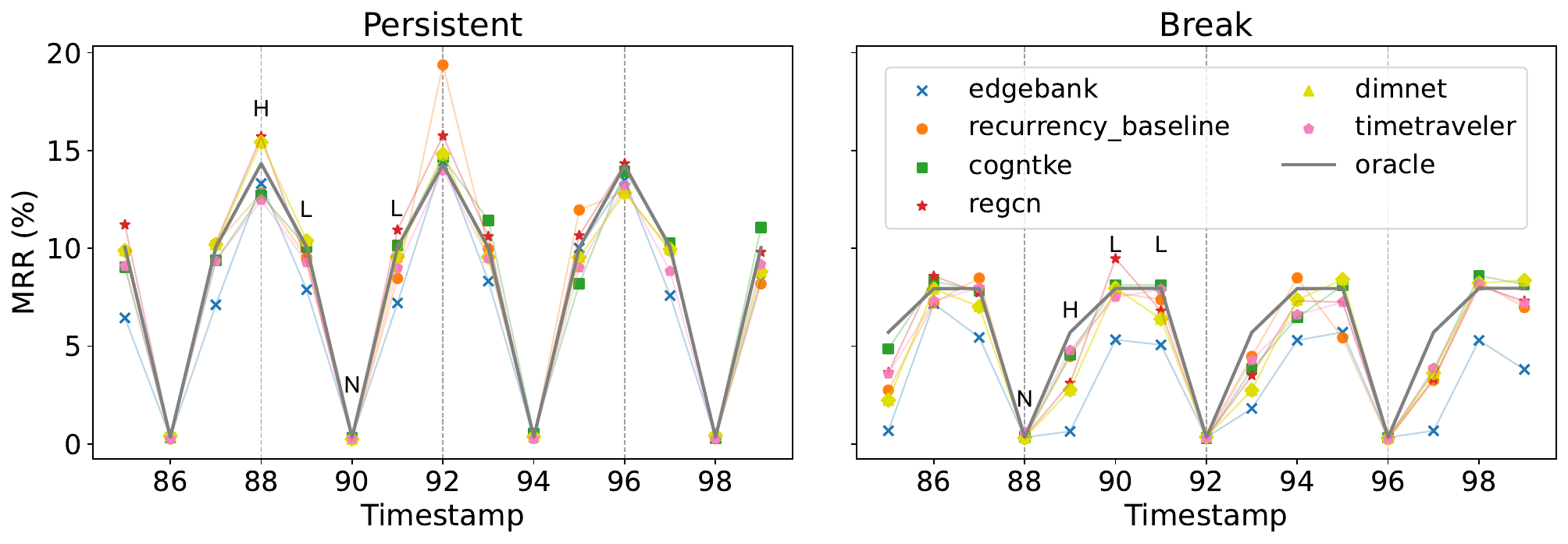}
\caption{MRR per test timestamp, averaged over five runs, for all methods and the oracle on the persistent (left) and break (right) periodicity variants. Dashed vertical lines indicate period-start. One period for each setting is annotated.}
\label{fig:mrrboth}
\end{figure}

The structural break changes the seasonal pattern as well as the signal-related subspace. It reduces the oracle reference from 8.30 to 5.83 MRR, which is predominantly due to the fact that both signal sources' subspaces expand after the break. Notably, all models lose oracle-normalised performance. While CognTKE achieves the strongest absolute break performance, EdgeBank-RX and TimeTraveler show the largest/smallest deterioration, respectively. 
To showcase performance across time, \autoref{fig:mrrboth} reports timestamp-level test MRR for the models in the persistent- and break setting, with the oracle depicted as a contiguous trajectory line. The figure primarily demonstrates that, across models, post-break performance follows the periodic structure more unevenly compared to the persistent setting. Moreover, periods indicated by the $L$ signal are recovered substantially better than those dominated by $H$\footnote{For the break setting, MRR scores grouped by signal can be found in~\autoref{tab:periodicity-break-pattern-results}.}.
Overall, our results suggest that periodicity is broadly learnable under stationary conditions, but that changes in seasonal support expose the limitations of methods relying on fact repetition.

\vspace{-0.3cm}
\vspace{-0.75cm}

\section{Conclusion and Final Remarks}
This section outlines takeaways, limitations, and directions for future research.

\mypar{Conclusion} 
Across our experiments, robustness proved highly signal-dependent: recurrence and periodicity were generally recoverable under stationary conditions, whereas structural breaks often led to substantial performance degradation. Homophily-driven structure posed the greatest challenge, suggesting that latent community dynamics remain difficult for current architectures to capture. Overall, our results indicate that benchmark performance alone provides only a partial view of model capabilities and that controlled analyses of underlying DGPs can yield complementary insights into what models actually learn.

\mypar{Limitations}  
Our conclusions are necessarily tied to the design choices of this study. Recurrence, homophily, and periodicity were instantiated through specific mechanisms and break configurations, while parameters such as break points and observation ratios were selected to provide controlled and interpretable settings. Consequently, our findings should be viewed as evidence of relative robustness within the studied scenarios rather than general statements about TKG models.

\mypar{Future Work} 
Motivated by insights from network science and time series analysis, this work constitutes an initial step toward a systematic study of distribution shifts in TKG forecasting. Future work could explore additional shift mechanisms, such as bursts of previously unseen entities, shifts in degree distributions as well as interactions thereof. Another important direction concerns the investigation of gradual shifts instead of abrupt changes. More broadly, we envision controlled synthetic benchmarks that complement real-world datasets and enable principled evaluation of robustness under evolving DGPs.



%
%
%
\bibliographystyle{splncs04}
\bibliography{bibliography}
\clearpage

\noindent
{\large\textbf{Appendix}}

\setcounter{table}{0}
\setcounter{figure}{0}
\setcounter{section}{0}

\renewcommand{\thesection}{A\arabic{section}}
\renewcommand{\thefigure}{A\arabic{figure}}
\renewcommand{\thetable}{A\arabic{table}}

\renewcommand{\theHsection}{appendix.section.\arabic{section}}
\renewcommand{\theHfigure}{appendix.figure.\arabic{figure}}
\renewcommand{\theHtable}{appendix.table.\arabic{table}}

\section{Model Parametrisations and Optimisation Regimes}\label{ap:models}
In the following, we provide an overview of the training and fine-tuning regimes used for the benchmark models. Hyperparameter tuning was performed as a grid search: candidate configurations were selected by validation MRR, while test evaluation was disabled during tuning and reserved for the final concrete model runs. Importantly, the hyperparameters are mainly selected from the ranges specified in the original papers and code repositories. Unless explicitly listed in the tuning grid in~\autoref{tab:HPO}, all model parameters were kept fixed at the defaults. 

\mypartwo{RE-GCN} RE-GCN was trained with the ConvTransE decoder and UVR-GCN encoder, using 200 hidden dimensions, 100 bases, dropout $0.2$, learning rate $10^{-3}$, gradient clipping at $1.0$, and early stopping with patience 20 over 50 epochs. Tuning varied the number of recurrent layers $n_{\text{layers}} \in \{1,2\}$ and the temporal history length $\in \{1,2,3,4,5,10,15\}$, applied jointly to train- and evaluation history.

\mypartwo{TLogic} TLogic was run as a rule-based baseline with minimum confidence $0.01$, minimum body support 2, and separate score parameters $\lambda$ and $\alpha$. The tuning grid varied the number of walks $\{10,100,200\}$, transition distribution $\{\mathrm{exp},\mathrm{unif}\}$, rule lengths $\{[1,2,3]\}$, temporal window $\{10,50,0\}$, top-$k \in \{10,20\}$, $\alpha \in \{0,0.5,1\}$, and $\lambda \in \{0.1,0.5,1\}$.

\mypartwo{Timetraveler} TimeTraveler used dynamic entity embeddings, path length 3, beam size 100, entity dimension 80, relation/state/hidden dimension 100, and time dimension 20. It was trained for up to 50 epochs with batch size 512, learning rate $10^{-3}$, gradient clipping at 10, and patience 20. Tuning varied only the maximum action number $\in \{30,50,60\}$.

\mypartwo{Recurrency baseline} The recurrency baseline used a window of 0 with both $\psi$ and $\xi$ scoring components enabled. The default fallback values were $\lambda_{\psi}=0.1$ and $\alpha=0.99999$. Its tuning regime activated learning of both $\lambda_{\psi}$ and $\alpha$ on the validation split, using the runner's internal grids, while leaving the rest of the scoring setup fixed.

\mypartwo{DimNet} DiMNet used 128 input dimensions, top-$k=50$, TransE messages, PNA aggregation, shortcut connections, layer normalisation, RReLU activation, and dropout $0.2$. It was trained for up to 50 epochs with learning rate $10^{-3}$, weight decay $10^{-4}$, gradient clipping at $1.0$, and patience 20. Tuning varied history length $\{10,2,5\}$, number of layers $\{1,3\}$, and number of attention heads $\{1,4\}$.

\mypartwo{CognTKE} CognTKE was run with the TRED\_GNN20 architecture, using 3 layers, a temporal window size of 10, hidden dimension 64, maximum global window size 5000, attention dimension 5, IDD activation, dropout $0.25$, time dimension 16, and regularisation parameter $\lambda=0.00012$. It was trained for up to 50 epochs with batch size 128, learning rate $5 \times 10^{-3}$, gradient clipping at $1.0$, and early stopping with patience 20. 

\noindent
Table~\ref{tab:HPO} details the selected hyperparameter configurations and their results.
\vspace{-0.7cm}
\begin{longtable}{>{\raggedright\arraybackslash}p{0.2\textwidth} >{\raggedright\arraybackslash}p{0.15\textwidth} p{0.2\textwidth} >{\raggedright\arraybackslash}p{0.4\textwidth}}
\caption{Validation-selected hyperparameter configurations used for experiments. Regime abbreviations: H-P/H-B = Homophily Persistent/Break, R-P/R-B = Recurrence Persistent/Break, P-P/P-B = Periodicity Persistent/Break.}\\
\toprule
Model & Regime & Val. MRR & Selected parameters \\
\midrule
\endfirsthead
\toprule
Model & Regime & Val. MRR & Selected parameters \\
\midrule
\endhead
\midrule
\multicolumn{4}{r}{Continued on next page} \\
\endfoot
\bottomrule
\endlastfoot
RE-GCN & H-P & 0.1679 & \texttt{his\_len=2, n\_ly=2} \\
RE-GCN & H-B & 0.0916 & \texttt{his\_len=1, n\_ly=2} \\
RE-GCN & R-P & 0.1601 & \texttt{his\_len=10, n\_ly=1} \\
RE-GCN & R-B & 0.1114 & \texttt{his\_len=15, n\_ly=1} \\
RE-GCN & P-P & 0.0929 & \texttt{his\_len=10, n\_ly=1} \\
RE-GCN & P-B & 0.0507 & \texttt{his\_len=4, n\_ly=2} \\
DiMNet & H-P & 0.2481 & \texttt{his\_len=5, n\_head=4, n\_ly=1} \\
DiMNet & H-B & 0.1237 & \texttt{his\_len=10, n\_head=4, n\_ly=1} \\
DiMNet & R-P & 0.1614 & \texttt{his\_len=2, n\_head=1, n\_ly=1} \\
DiMNet & R-B & 0.1110 & \texttt{his\_len=5, n\_head=1, n\_ly=3} \\
DiMNet & P-P & 0.0950 & \texttt{his\_len=10, n\_head=1, n\_ly=1} \\
DiMNet & P-B & 0.0497 & \texttt{his\_len=5, n\_head=4, n\_ly=1} \\
TimeT & H-P & 0.2214 & \texttt{max\_action\_n=30} \\
TimeT & H-B & 0.1239 & \texttt{max\_action\_n=30} \\
TimeT & R-P & 0.1661 & \texttt{max\_action\_n=50} \\
TimeT & R-B & 0.1209 & \texttt{max\_action\_n=50} \\
TimeT & P-P & 0.0894 & \texttt{max\_action\_n=50} \\
TimeT & P-B & 0.0499 & \texttt{max\_action\_n=30} \\
TLogic & H-P & 0.0954 & \texttt{$\alpha$=0, $\lambda$=0.1, n\_walks=200, top\_k=10, t\_distr=exp, win=0} \\
TLogic & H-B & 0.0461 & \texttt{$\alpha$=0.5, $\lambda$=0.5, n\_walks=200, top\_k=10, t\_distr=unif, win=0} \\
TLogic & R-P & 0.1604 & \texttt{$\alpha$=0, $\lambda$=0.1, n\_walks=200, top\_k=10, t\_distr=unif, win=10} \\
TLogic & R-B & 0.1147 & \texttt{$\alpha$=0, $\lambda$=0.1, n\_walks=10, top\_k=10, t\_distr=exp, win=10} \\
TLogic & P-P & 0.0930 & \texttt{$\alpha$=1, $\lambda$=0.1, n\_walks=200, top\_k=20, t\_distr=unif, win=50} \\
TLogic & P-B & 0.0462 & \texttt{$\alpha$=1, $\lambda$=0.1, n\_walks=200, top\_k=20, t\_distr=exp, win=50} \\
RecBL & H-P & 0.0187 & \texttt{learn\_$\alpha$=true, learn\_$\lambda$\_psi=true} \\
RecBL & H-B & 0.0103 & \texttt{learn\_$\alpha$=true, learn\_$\lambda$\_psi=true} \\
RecBL & R-P & 0.1699 & \texttt{learn\_$\alpha$=true, learn\_$\lambda$\_psi=true} \\
RecBL & R-B & 0.1240 & \texttt{learn\_$\alpha$=true, learn\_$\lambda$\_psi=true} \\
RecBL & P-P & 0.0995 & \texttt{learn\_$\alpha$=true, learn\_$\lambda$\_psi=true} \\
RecBL & P-B & 0.0555 & \texttt{learn\_$\alpha$=true, learn\_$\lambda$\_psi=true} 
\label{tab:HPO}
\end{longtable}

\vspace{-0.7cm}


\section{Self-referential Facts}\label{sec:self_ref_facts}
As indicated via~\autoref{tab:self-referential-facts}, facts where the subject- and object entity coincide tend to occur quite rarely and, at times, not at all. For this reason, we make the simplifying assumption of excluding such observations throughout our experiments for representative- and ease-of-modelling purposes. The incorporation of such facts may of course be pursued by the interested reader.

\vspace{-0.7cm}
\begin{table}[h!]
\centering
\caption{Self-referential facts across real temporal knowledge graph datasets.}
\label{tab:self-referential-facts}
\setlength{\tabcolsep}{4pt}
\begin{tabular}{lrrr}
\toprule
Dataset & $\#\textnormal{obs}$ & Self-ref. facts & Self-ref. pct \\
\midrule
ICEWS18 & 468,558 &  0 & $0.0000\%$ \\
ICEWS14 &  90,730 &  1 & $0.0011\%$ \\
WIKI    & 669,934 & 79 & $0.0118\%$ \\
YAGO    & 201,089 &  0 & $0.0000\%$ \\
\midrule
MEAN    & 357,577.75 & 20 & $0.0032\%$ \\
\bottomrule
\end{tabular}
\end{table}
\vspace{-0.7cm}

\section{Observation Ratio Groundwork}\label{sec:observation_ratio}
As a guardrail for our data generation mechanism, we leverage an \emph{observation ratio}, which we lay out as the ratio of seen vs. possible facts across all timestamps. In principle, we assume data to be non-self-referential and calculate the size of the induced universe of fact triplets via $|\Omega| = n_r(n_e^2 -n_e)$. Dividing the number of observations of a real dataset across all timestamps by this factor yields a heuristic-type estimate for our synthetic data size. This estimate, in combination with $n_e,n_r$ being set a-priori, allows for the calculation of the number of observations relative to the fact universe. For example, in YAGO, $|\Omega|$ amounts to $1.128\times 10^9$. The number of observations in the dataset lies at $201089$, yielding an observation ratio of $1.78211 \times 10^{-4}$. The approximate mean ratio over a set of well-established TKGs (cf.~\cite{REGCN} for base statistics) is approx. $10^{-4}$, as shown in Table~\ref{tab:dataset-statistics}. Therefore, in this work, we set the total number of observations for a dataset as the product of this ratio with the size of the fact universe $\vert \Omega \vert$.

\vspace{-0.7cm}
\begin{table}[h!]
\centering
\caption{Dataset characteristics and observation ratio.}
\label{tab:dataset-statistics}
\setlength{\tabcolsep}{4pt}
\begin{tabular}{lrrrrr}
\toprule
Dataset & $\#\textnormal{obs}$ & $n_e$ & $n_r$ & $|\Omega|$ & Obs. ratio \\
\midrule
ICEWS18    &   468,558 & 23,033 & 256 & 135,806,990,336 & $3.45018 \times 10^{-6}$ \\
ICEWS14    &    90,730 &  6,869 & 230 &  10,850,547,160 & $8.36179 \times 10^{-6}$ \\
ICEWS05-15 &   461,329 & 10,094 & 251 &  25,571,564,242 & $1.80407 \times 10^{-5}$ \\
WIKI       &   669,934 & 12,554 &  24 &   3,782,168,688 & $1.77130 \times 10^{-4}$ \\
YAGO       &   201,089 & 10,623 &  10 &   1,128,375,060 & $1.78211 \times 10^{-4}$ \\
GDELT      & 2,278,405 &  7,691 & 240 &  14,194,509,600 & $1.60513 \times 10^{-4}$ \\
\midrule
MEAN       &   695,007.5 & 11,810.7 & 168.5 & 31,889,025,847.7 & $9.09511 \times 10^{-5}$ \\
\bottomrule
\end{tabular}
\end{table}

\vspace{-0.7cm}

\section{On Hyperparameters for the TKG Generator}\label{ap:hyperparam_choices_synTKG}
The hyperparameters of each data variant affect the experiments in two related ways: they determine the strength of the intended signal and, within this signal regime, the resulting forecasting difficulty. In the recurrence variant, for example, increasing $\rho$ makes historical facts dominate the sampling distribution. In the limit $\rho \rightarrow \infty$, this leads to near-deterministic resampling and therefore to a trivial forecasting task, with MRR values close to $100\%$. Conversely, as $\rho \rightarrow 0$, the recurrence signal vanishes. Hence, useful parameter choices lie in an admissible range in $\mathbb{R}^+$ where signal strength and forecasting difficulty remain balanced. Preliminary experiments indicated that $\rho = 7.5$ provides a suitable representative of this range. The hyperparameters of the remaining simulation variants were chosen according to the same principle. Since the results depend on these choices, we aim at systematic analyses regarding the sensitivity of both the generated data distributions and model performance to the selected parameters for future work.

\section{Additional Results}
Here, we provide additional results that accompany our main paper. Table~\ref{tab:periodicity-break-pattern-results} shows the MRR results for the periodicity dataset under the break variant, separated for the two signals H and L. Further, Tables~\ref{tab:periodicity-hits-results}--\ref{tab:recurrence-hits-results} show the Hits@k results, and Table~\ref{tab:runtime-all-variants} reports the runtimes. Experiments were run on a system encompassing an NVIDIA A40 (48GB VRAM) and an AMD EPYC 7713P (2GHz@64 Cores, 128 Threads) with 512GB of RAM.


\begin{table}[h!]
\centering
\caption{Ablation -- Periodicity, break variant: test MRR by active signal pattern. Scores are mean (standard deviation) over five runs, in percent.}
\label{tab:periodicity-break-pattern-results}
\setlength{\tabcolsep}{3pt}
\begin{tabular}{lrrrr} 
\toprule
Model & H MRR (s.d.) & H \%Oracle & L MRR (s.d.) & L \%Oracle \\ 
\midrule
\textsc{Oracle} & 5.71 (0.00) & 100.00 & 7.95 (0.00) & 100.00 \\ 
\midrule
CognTKE & \underline{3.93} (0.33) & 68.83 & \textbf{7.95} (0.40) & 100.00 \\ 
DiMNet & 2.99 (0.39) & 52.36 & 7.47 (0.17) & 93.96 \\ 
EdgeBank-RX & 0.95 (0.00) & 16.64 & 5.39 (0.00) & 67.80 \\ 
RecurrencyBaseline & 3.73 (0.00) & 65.32 & 7.50 (0.00) & 94.34 \\ 
RE-GCN & 3.71 (0.98) & 64.97 & \underline{7.81} (0.17) & 98.24 \\ 
TimeTraveler & \textbf{4.13} (0.52) & 72.33 & 7.49 (0.47) & 94.21 \\ 
TLogic & 1.43 (0.11) & 25.04 & 7.46 (0.20) & 93.84 \\ 
\bottomrule
\end{tabular}
\end{table}

\begin{table}[t]
\centering
\caption{Runtime across all data-variant and effect-mode combinations. Scores are mean (s.d.) in minutes. p. indicates persistent, b. indicates break.}
\label{tab:runtime-all-variants}
\scriptsize
\setlength{\tabcolsep}{2.5pt}
\begin{tabular}{lrrrrrr}
\toprule
Model & Period.\ p. & Period.\ b. & Homoph.\ p. & Homoph.\ b. & Recur.\ p. & Recur.\ b. \\
\midrule
CognTKE & 4.46 (1.78) & 3.02 (0.53) & 2.58 (0.64) & 2.99 (1.51) & 2.56 (0.58) & 3.12 (0.44) \\
DiMNet & 3.80 (0.54) & 2.05 (0.30) & 2.91 (0.55) & 5.05 (0.71) & 1.56 (0.36) & 3.32 (1.36) \\
EdgeBank & 0.08 (0.00) & 0.08 (0.00) & 0.08 (0.00) & 0.09 (0.01) & 0.08 (0.00) & 0.09 (0.01) \\
RecBL & 5.31 (0.03) & 5.29 (0.01) & 6.36 (0.02) & 6.94 (0.80) & 6.77 (0.74) & 6.22 (0.01) \\
RE-GCN & 4.57 (0.93) & 1.92 (0.07) & 2.53 (0.04) & 1.87 (0.03) & 4.04 (0.84) & 7.05 (1.00) \\
TimeT & 31.33 (8.78) & 19.90 (3.73) & 4.25 (1.32) & 3.82 (0.71) & 13.91 (3.09) & 13.18 (3.13) \\
TLogic & 8.36 (1.01) & 3.52 (0.08) & 0.30 (0.01) & 0.30 (0.05) & 0.09 (0.00) & 0.07 (0.00) \\
\bottomrule
\end{tabular}
\end{table}

\clearpage

\begin{table}[t]
\centering
\caption{Hits@k performance on the periodicity variants. Scores are mean (s.d.) over five runs, in percent; \%O gives the percentage of the corresp. oracle score.}
\label{tab:periodicity-hits-results}
\scriptsize
\setlength{\tabcolsep}{2pt}
\renewcommand{\arraystretch}{0.92}
\resizebox{\textwidth}{!}{%
\begin{tabular}{@{}l*{12}{c}@{}}
\toprule
& \multicolumn{6}{c}{Persistent} & \multicolumn{6}{c}{Break} \\
\cmidrule(lr){2-7}\cmidrule(lr){8-13}
Model & H@1 & \%O & H@3 & \%O & H@10 & \%O & H@1 & \%O & H@3 & \%O & H@10 & \%O \\
\midrule
Oracle & 1.97 (0.00) & 100.00 & 5.91 (0.00) & 100.00 & 19.71 (0.00) & 100.00 & 1.22 (0.00) & 100.00 & 3.66 (0.00) & 100.00 & 12.19 (0.00) & 100.00 \\
Random & 0.04 (0.00) & 2.03 & 0.12 (0.00) & 2.03 & 0.40 (0.00) & 2.03 & 0.04 (0.00) & 3.28 & 0.12 (0.00) & 3.28 & 0.40 (0.00) & 3.28 \\
\midrule
CognTKE & 1.91 (0.29) & 96.77 & 5.95 (0.15) & 100.64 & 19.77 (0.32) & 100.29 & 1.21 (0.14) & 98.96 & 3.40 (0.36) & 93.04 & 11.34 (0.78) & 93.01 \\
DiMNet & 1.98 (0.13) & 100.37 & 5.80 (0.38) & 98.11 & 19.51 (0.42) & 98.94 & 0.82 (0.08) & 67.46 & 2.87 (0.32) & 78.62 & 10.60 (0.55) & 86.99 \\
EdgeBank & 1.94 (0.00) & 98.20 & 5.81 (0.00) & 98.22 & 19.35 (0.00) & 98.16 & 0.99 (0.00) & 81.15 & 2.81 (0.00) & 76.83 & 8.21 (0.00) & 67.37 \\
Recurrency & 2.06 (0.00) & 104.27 & 5.64 (0.00) & 95.31 & 19.75 (0.00) & 100.20 & 1.06 (0.00) & 87.15 & 3.29 (0.00) & 90.04 & 10.57 (0.00) & 86.73 \\
RE-GCN & 1.94 (0.28) & 98.33 & 5.96 (0.55) & 100.82 & 19.61 (0.53) & 99.49 & 1.16 (0.22) & 94.89 & 3.60 (0.19) & 98.54 & 11.95 (0.74) & 98.01 \\
TimeTraveler & 1.66 (0.27) & 84.09 & 5.35 (0.43) & 90.55 & 18.44 (0.47) & 93.54 & 1.24 (0.25) & 102.02 & 3.37 (0.24) & 92.10 & 10.84 (0.56) & 88.91 \\
TLogic & 1.68 (0.22) & 	85.33 & 5.14 (0.41) & 86.94 & 18.11 (0.43) & 91.87 & 0.86 (0.13) & 70.94 & 2.77 (0.13) & 75.87 & 9.82 (0.29) & 80.58 \\
\bottomrule
\end{tabular}%
}
\end{table}

\begin{table}[t]
\centering
\caption{Hits@k performance on the homophily variants. Scores are mean (s.d.) over five runs, in percent; \%O gives the percentage of the corresp. oracle score.}
\label{tab:homophily-hits-results}
\scriptsize
\setlength{\tabcolsep}{2pt}
\renewcommand{\arraystretch}{0.92}
\resizebox{\textwidth}{!}{%
\begin{tabular}{@{}l*{12}{c}@{}}
\toprule
& \multicolumn{6}{c}{Persistent} & \multicolumn{6}{c}{Break} \\
\cmidrule(lr){2-7}\cmidrule(lr){8-13}
Model & H@1 & \%O & H@3 & \%O & H@10 & \%O & H@1 & \%O & H@3 & \%O & H@10 & \%O \\
\midrule
Oracle & 9.67 (0.00) & 100.00 & 29.00 (0.00) & 100.00 & 86.97 (0.00) & 100.00 & 9.63 (0.00) & 100.00 & 28.89 (0.00) & 100.00 & 86.64 (0.00) & 100.00 \\
Random & 0.04 (0.00) & 0.41 & 0.12 (0.00) & 0.41 & 0.40 (0.00) & 0.46 & 0.04 (0.00) & 0.42 & 0.12 (0.00) & 0.42 & 0.40 (0.00) & 0.46 \\
\midrule
CognTKE & 8.70 (0.55) & 89.97 & 25.66 (0.51) & 88.50 & 56.53 (0.21) & 65.00 & 6.75 (0.21) & 70.12 & 18.31 (0.49) & 63.37 & 41.31 (0.30) & 47.68 \\
DiMNet & 14.27 (0.88) & 147.65 & 28.16 (0.93) & 97.10 & 43.40 (1.73) & 49.90 & 7.53 (0.39) & 78.17 & 15.11 (0.78) & 52.30 & 23.88 (0.79) & 27.56 \\
EdgeBank & 1.33 (0.00) & 13.76 & 1.58 (0.00) & 5.45 & 1.86 (0.00) & 2.13 & 1.29 (0.00) & 13.43 & 1.57 (0.00) & 5.44 & 1.86 (0.00) & 2.14 \\
Recurrency & 0.97 (0.00) & 10.04 & 1.22 (0.00) & 4.19 & 1.43 (0.00) & 1.65 & 1.02 (0.00) & 10.62 & 1.24 (0.00) & 4.28 & 1.45 (0.00) & 1.68 \\
RE-GCN & 8.40 (0.26) & 86.93 & 20.92 (0.65) & 72.15 & 36.57 (0.99) & 42.05 & 4.75 (0.21) & 49.36 & 11.96 (0.28) & 41.40 & 20.39 (0.50) & 23.53 \\
TimeTraveler & 8.84 (0.44) & 91.50 & 24.12 (1.46) & 83.20 & 62.34 (1.85) & 71.67 & 5.41 (0.56) & 56.23 & 15.88 (1.33) & 54.98 & 39.68 (2.92) & 45.80 \\
TLogic & 7.10 (0.58) & 73.44 & 14.25 (0.42) & 49.15 & 15.86 (0.53) & 18.23 & 4.14 (0.21) & 43.01 & 7.87 (0.33) & 27.24 & 8.83 (0.54) & 10.19 \\
\bottomrule
\end{tabular}%
}
\end{table}

\begin{table}[t]
\centering
\caption{Hits@k performance on the recurrence variants. Scores are mean (s.d.) over five runs, in percent; \%O gives the percentage of the corresp. oracle score.}
\label{tab:recurrence-hits-results}
\scriptsize
\setlength{\tabcolsep}{2pt}
\renewcommand{\arraystretch}{0.92}
\resizebox{\textwidth}{!}{%
\begin{tabular}{@{}l*{12}{c}@{}}
\toprule
& \multicolumn{6}{c}{Persistent} & \multicolumn{6}{c}{Break} \\
\cmidrule(lr){2-7}\cmidrule(lr){8-13}
Model & H@1 & \%O & H@3 & \%O & H@10 & \%O & H@1 & \%O & H@3 & \%O & H@10 & \%O \\
\midrule
Oracle & 20.01 (0.00) & 100.00 & 20.08 (0.00) & 100.00 & 20.30 (0.00) & 100.00 & 16.44 (0.00) & 100.00 & 16.54 (0.00) & 100.00 & 16.77 (0.00) & 100.00 \\
Random & 0.04 (0.00) & 0.20 & 0.12 (0.00) & 0.60 & 0.40 (0.00) & 1.97 & 0.04 (0.00) & 0.24 & 0.12 (0.00) & 0.73 & 0.40 (0.00) & 2.39 \\
\midrule
CognTKE & 20.00 (0.02) & 99.95 & 20.05 (0.04) & 99.87 & 20.25 (0.06) & 99.77 & 16.25 (0.23) & 98.85 & 16.38 (0.23) & 99.04 & 16.62 (0.21) & 99.08 \\
DiMNet & 15.23 (0.22) & 76.12 & 15.86 (0.18) & 79.01 & 17.13 (0.53) & 84.36 & 10.82 (0.22) & 65.79 & 12.98 (0.65) & 78.48 & 14.09 (0.73) & 84.00 \\
EdgeBank & 17.70 (0.00) & 88.44 & 19.95 (0.00) & 99.38 & 20.30 (0.00) & 100.00 & 14.69 (0.00) & 89.38 & 16.54 (0.00) & 99.98 & 16.77 (0.00) & 99.98 \\
Recurrency & 20.10 (0.00) & 100.41 & 20.26 (0.00) & 100.89 & 20.50 (0.00) & 101.00 & 16.04 (0.00) & 97.59 & 16.20 (0.00) & 97.96 & 16.50 (0.00) & 98.34 \\
RE-GCN & 16.95 (0.32) & 84.70 & 19.39 (0.36) & 96.59 & 19.83 (0.45) & 97.70 & 12.72 (1.05) & 77.38 & 14.56 (0.61) & 88.02 & 15.46 (0.52) & 92.17 \\
TimeTraveler & 18.94 (0.95) & 94.62 & 19.83 (0.32) & 98.75 & 20.23 (0.05) & 99.65 & 14.78 (0.66) & 89.90 & 16.26 (0.21) & 98.33 & 16.71 (0.14) & 99.60 \\
TLogic & 18.70 (0.27) & 93.43 & 18.78 (0.27) & 93.55 & 19.01 (0.27) & 93.63 & 12.67 (0.69) & 77.06 & 12.77 (0.68) & 77.22 & 13.02 (0.68) & 77.60 \\
\bottomrule
\end{tabular}%
}
\end{table}


\end{document}